\documentclass[letterpaper, 10 pt, conference]{ieeeconf}  

\IEEEoverridecommandlockouts                              

\usepackage{epsfig} 
\usepackage{amsmath} 
\usepackage{amssymb}  
\usepackage{booktabs}
\usepackage[dvipsnames]{xcolor}
\usepackage{url}
\usepackage{caption} 
\usepackage{graphicx}
\usepackage{url}
\usepackage{cite}

\usepackage{setspace}

\renewcommand{\baselinestretch}{0.96}

\title{\LARGE \textbf{Periodic Steady-State Control of a Handkerchief-Spinning Task Using a Parallel Anti-Parallelogram Tendon-driven Wrist}
}

\author{Lei Liu$^{1,2,*}$, Haonan Zhang$^{1,3,*}$, Huahang Xu$^{4}$, Zefan Zhang$^{1,5}$, Lulu Chang$^{1,6}$, Lei Lv$^{1,7}$\\Andrew Ross McIntosh$^{8}$, Kai Sun$^{1,2}$, Zhenshan Bing$^{9,10}$, Jiahong Dong$^{11,12,13\dag}$, Fuchun Sun$^{1,\dag}$,~\IEEEmembership{Fellow,~IEEE}\thanks{*These authors contributed equally to this research (E-Mail:\protect\url{liul22@mails.tsinghua.edu.cn}). $^{\dag}$Co-corresponding authors: Jiahong Dong and Fuchun Sun (\protect\url{fcsun@mail.tsinghua.edu.cn}). $^{1}$Department of Computer Science and Technology, Tsinghua University, Beijing, China. $^{2}$School of Biomedical Engineering, Tsinghua University, Beijing, China. $^{3}$School of Artificial Intelligence, Beihang University, Beijing, China. $^{4}$Institute of Nuclear and New Energy Technology, Tsinghua University, Beijing, China. $^{5}$Tsinghua Shenzhen International Graduate School, Tsinghua University, Shenzhen, China. $^{6}$School of Automation, Nanjing University of Science and Technology, Nanjing, China. $^{7}$Shanghai Research Institute for Intelligent Autonomous Systems, Tongji University, Shanghai, China. $^{8}$Department of Mechanical Engineering, Tsinghua University, Beijing, China. $^{9}$School of Computation, Information and Technology, Technical University of Munich, Garching, Germany. $^{10}$State Key Laboratory for Novel Software Technology and the School of Science and Technology, Nanjing University (Suzhou Campus), Suzhou, China. $^{11}$Hepato-pancreato-biliary Center, Beijing Tsinghua Changgung Hospital, Beijing, China. $^{12}$Key Laboratory of Digital Intelligence Hepatology (Ministry of Education), Beijing, China. $^{13}$School of Clinical Medicine, Tsinghua Medicine, Tsinghua University, Beijing, China.}}

\begin{document}

\maketitle

\thispagestyle{empty}
\pagestyle{empty}

\begin{abstract}
Spinning flexible objects, exemplified by traditional Chinese handkerchief performances, demands periodic steady-state motions under nonlinear dynamics with frictional contacts and boundary constraints. To address these challenges, we first design an intuitive dexterous wrist based on a parallel anti-parallelogram tendon-driven structure, which achieves $90^\circ$ omnidirectional rotation with low inertia and decoupled roll–pitch sensing, and implement a high–low level hierarchical control scheme. We then develop a particle–spring model of the handkerchief for control-oriented abstraction and strategy evaluation. Hardware experiments validate this framework, achieving an unfolding ratio of approximately $99\%$ and fingertip tracking error of $\mathrm{RMSE}=2.88\,\text{mm}$ in high-dynamic spinning. These results demonstrate that integrating control-oriented modeling with a task-tailored dexterous wrist enables robust rest-to–steady-state transitions and precise periodic manipulation of highly flexible objects. More visualizations: \textcolor{Cerulean}{\protect\url{https://slowly1113.github.io/icra2026-handkerchief/}}.
\end{abstract}

\section{INTRODUCTION}

Dynamic manipulation of flexible objects is an increasingly important yet challenging problem in robotics. In this work, we focus on the traditional Chinese task of \emph{handkerchief spinning}, where a square fabric is driven into a stable periodic rotation by dexterous wrist actions. While significant progress has been made in the static and quasi-static handling of rigid \cite{yao2025soft} or quasi-rigid \cite{sundaralingam2018geometric} deformables and highly flexible objects, dynamic control of highly flexible objects \cite{yoo2024moe,nocentini2022learning} remains in its infancy.

From a hardware perspective, the human wrist comprises a parallel array of carpal bones actuated by antagonistic musculotendinous units with rich sensorimotor feedback \cite{behnke2012kinetic}. This skeletal–musculotendinous–neural architecture combines compliance with load-bearing capacity, yielding low effective inertia and high dynamic responsiveness \cite{levin2002tensegrity}. By contrast, conventional robotic wrists implemented as serial motor stacks concentrate inertia and exhibit limited intrinsic dexterity. They often require additional degrees of freedom and sophisticated motion planning to approximate tasks such as \emph{handkerchief spinning}, which reduces real-time responsiveness and increases control complexity \cite{fan2022prosthetic}. These observations motivate rethinking the wrist mechanism itself. Recent studies have therefore explored parallel wrist architectures to improve dynamic response \cite{kim2018quaternion,zhengNovelVariableStiffness2025,xieWristinspiredSuspendedTubercletype2023}, partly because such parallel, tendon-driven layouts are more similar to the human skeletal–muscular–neural system\cite{levin2002tensegrity}. While promising, these designs still struggle to balance precision and dexterity, limiting their ability to support uniform axial spinning in highly dynamic tasks \cite{shahComparisonRobotWrist2019}.

\begin{figure}[!t]
\centering
\includegraphics[width=0.48\textwidth]{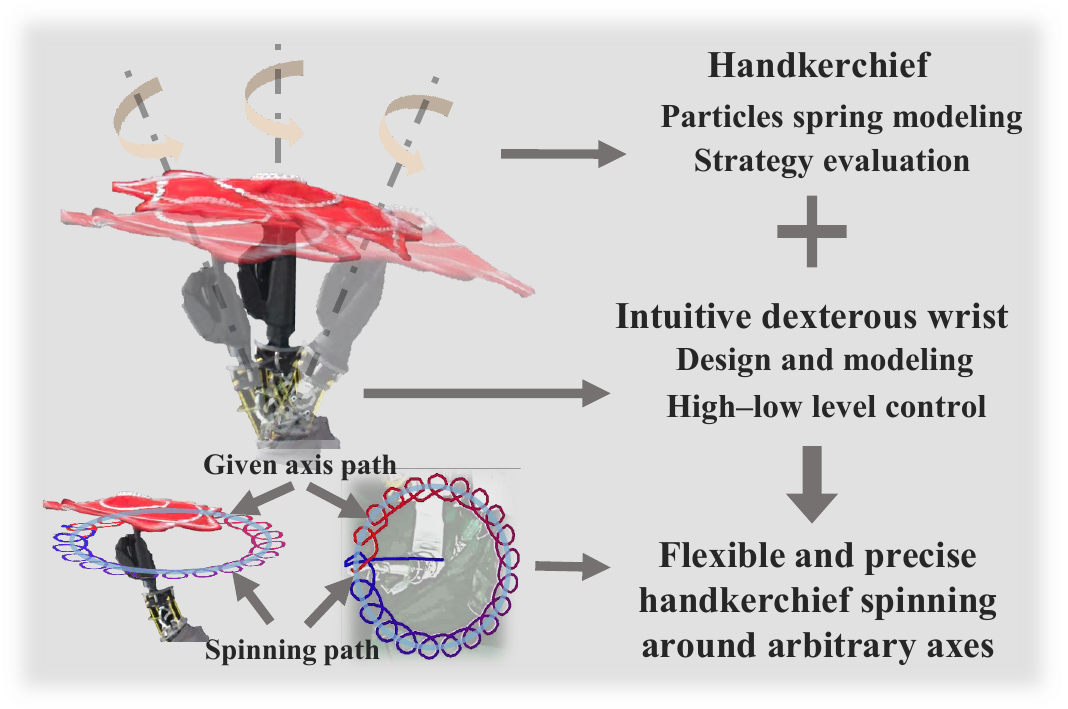}
\caption{Overview of the framework: particle–spring modeling for strategy evaluation, combined with a tendon-driven dexterous wrist and high–low level control, enabling flexible and precise handkerchief spinning around arbitrary axes.}
\label{fig_abstract}
\end{figure}

Humans can intuitively inject energy and exploit friction to initiate and sustain fabric spinning under boundary constraints. For robots, however, this process is governed by frictional contact, inertia, and nonlinear elasticity, making it highly nonlinear, difficult to model, and challenging to reproduce in a stable manner. Existing research can be broadly categorized into three directions: \textbf{(i)} tasks such as folding, spreading, or arranging cloths \cite{zhou2025dual,ganapathi2021learning}; \textbf{(ii)} approaches that rely on object inertia without explicitly modeling fabric dynamics \cite{chen2022efficiently}; and \textbf{(iii)} graphics-based simulations that emphasize visual realism \cite{mozafary2016study,magnenat2007measured}, while fiber–fiber contact models remain difficult to incorporate into controller design \cite{durville2010simulation}. In robotics, studies on highly dynamic deformable object manipulation are primarily confined to one-dimensional settings \cite{han2018towards,zhang2021robots}. Consequently, although perception and planning methods can alleviate geometric uncertainty, they fall short in addressing dynamic regimes dominated by frictional interactions and nonlinear behavior. From a control perspective, the initiation and stabilization of friction-driven fabric spinning remain largely unexplored, underscoring the need for an integrated approach that unifies dedicated hardware design with tractable control strategies.

\begin{figure*}[!t]
\centering
\includegraphics[width=1\textwidth]{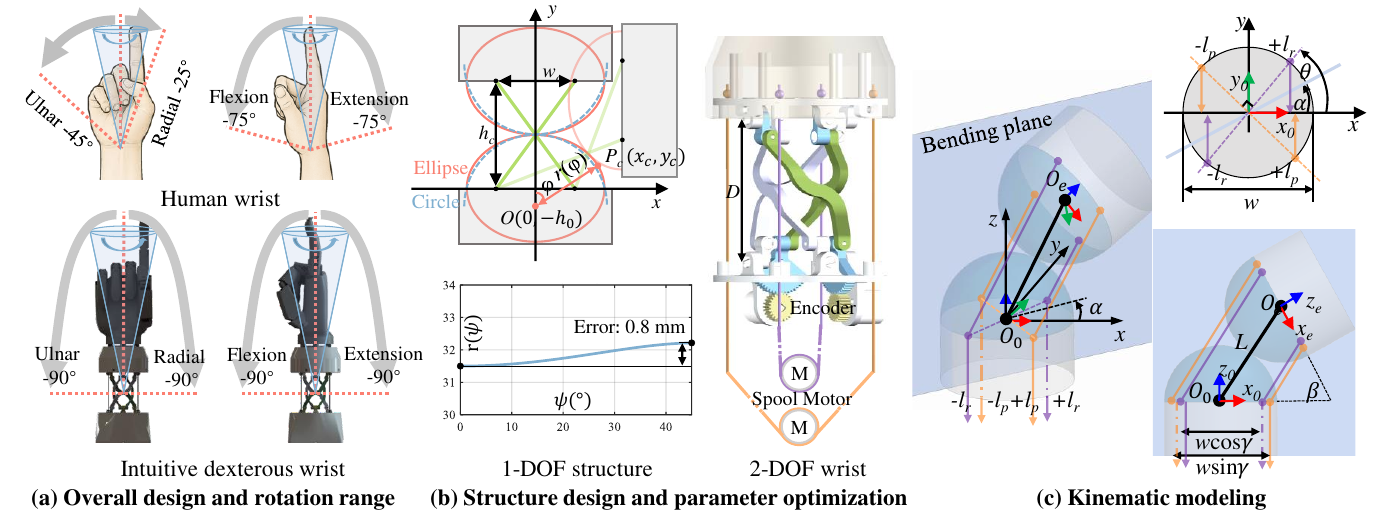}
\caption{Overview of the dexterous wrist and its kinematics. \textbf{(a)}: Parallel anti-parallelogram tendon-driven wrist architecture using re-parameterized anti-parallelogram structures to shorten $D$, reduce footprint/inertia, and enable $\sim90^{\circ}$ omnidirectional rotation; rear-mounted motors with tendon guides minimize distal inertia, and two decoupled encoders provide precise roll–pitch sensing. \textbf{(b)}: Design details and parameter optimization showing a nearly constant rolling radius $r(\psi)$; stacking two orthogonal modules yields a compact 2-DOF wrist. \textbf{(c)}: Kinematic abstraction as a spherical rolling-contact joint, with coordinate frames and tendon fixation indicated, and the corresponding tendon-length variation in the bending plane.}
\label{section2}
\end{figure*}
In this work, we adopt a systematic perspective, developing the dexterous wrist (Fig.~1) to equip a robot hand with the dexterity and high-dynamic capability required to spin a handkerchief. Then we model the initiation period and the steady spinning period. Our core contributions are as follows:

\begin{itemize}
  \item \textbf{Parallel anti-parallelogram tendon-driven wrist and hierarchical control.}
  We design a tendon-driven wrist with parallel anti-parallelogram links that combines high dynamic performance with dexterity. The structure achieves \(90^\circ\) omnidirectional rotation while maintaining low inertia. A task-tailored two-level controller regulates initiation with kinematic control. Once steady spinning emerges, it switches to direct periodic motor commands to sustain the motion with improved stability.

  \item \textbf{Control-oriented modeling and initiation strategy.}
  We develop a control-oriented initiation model and strategies that yield robust transitions from rest to steady spinning, relying on periodicity, energy balance, and phase locking rather than explicit friction models.

  \item \textbf{Empirical validation and key findings.}
  Through simulations and hardware experiments, we demonstrate low-inertia, high-dynamic axial rotations and robust rest-to–steady-state transitions. In the periodic regime, direct motor-level periodic commands sustain the spin and enhance dynamic performance and stability, validating the proposed initiation strategy and hierarchical control.
\end{itemize}

\section{Design and modeling of the Parallel Anti-Parallelogram Tendon-driven Wrist}


\subsection{Overview Design of the Parallel Anti-Parallelogram Tendon-driven Wrist }

Guided by wrist anatomy~\cite{bajajStateArtArtificial2019}, we adopt a tendon-driven dexterous wrist with parallel links. Our design builds upon the Quaternion Joint~\cite{kim2018quaternion} with the following \emph{modifications}: \textbf{(i)} Compared with the original Quaternion Joint, our design reduces the length-to-width ratio by about $50\%$, thereby decreasing both the overall footprint and the effective inertia.  
\textbf{(ii) }We further redesign the parallel link structure to preserve omnidirectional $90^\circ$ rotation (Fig.~2(a)), extending the mobility ranges of the human wrist (about $45^\circ$ ulnar, $25^\circ$ radial, and $75^\circ$ flexion/extension~\cite{bajajStateArtArtificial2019}) by approximately $100\%$, $260\%$, and $20\%$, respectively. In addition, two decoupled joint-angle encoders (one per orthogonal subchain) are integrated to provide precise roll–pitch sensing without cross-coupling.  
\textbf{(iii)} A task-tailored two-level hierarchical control is adopted: during initiation, kinematic regulation ensures a reliable rest-to–steady-state transition, and once stable spin is reached, the controller simplifies to direct periodic motor commands for efficient maintenance, leveraging the joint’s decoupled sensing advantage and high dynamic responsiveness.

\subsection{Detailed Design of the Anti-parallelogram Parallel Structure}
Fig.~2(b) shows the structural schematic of the proposed parallel anti-parallelogram tendon-driven wrist. An anti-parallelogram mechanism generates an ellipse (red curve) that approximates the motion of a circular rolling surface (blue curve). Here, we aim to shorten the wrist length $D$ to reduce the space occupied by the wrist and simultaneously decrease its inertia.

For the red curve in Fig.~2(b), which represents the actual elliptical trajectory formed by the link, the mathematical expression can be written as:
\begin{equation}
\label{1}
\frac{x_c^2}{(l_c/2)^2}+\frac{y_c^2}{(h_c/2)^2}=1.
\end{equation}

For this anti-parallelogram structure, the center $O$ of the approximated circular rolling surface and its distance to a point $P_c$ on the ellipse can be expressed as:
\begin{equation}
\label{2}
\left\{
\begin{aligned}
x_c &= r(\psi)\,\sin\psi,\\
y_c &= \frac{1}{\tan\psi}\,x_c - h_o,
\end{aligned}
\right.
\end{equation}
where $\psi$ denotes the angle between the connecting line and the $y$-axis.

Combining Eq.~\eqref{1} and Eq.~\eqref{2}, the distance $r(\psi)$ from the circle center $O$ to point $P_c$ on the ellipse can be expressed as a function of $\psi$:
\begin{equation}
\label{3}
r(\psi)=\frac{\,h_o+\sqrt{\,h_o^2+\left(1+\frac{h_c^2}{h_c^2+w_c^2}\tan^2\psi\right)\left(\left(\frac{h_c}{2}\right)^2-h_o^2\right)}\,}{\;\cos\psi\left(1+\frac{h_c^2}{h_c^2+w_c^2}\tan^2\psi\right)}.
\end{equation}

Using Eq.~\eqref{3}, appropriate $(h_o,\,w_c,\,h_c)$ values can be selected. If $r(\psi)$ remains nearly constant, the structure approximates a circular rolling surface. With the parameters $h_o=8\,\mathrm{mm}$, $w_c=35\,\mathrm{mm}$, and $h_c=47\,\mathrm{mm}$, where $\psi$ ranges from $0^\circ$ to $45^\circ$ (corresponding to actual wrist rotation of $-90^\circ$ to $90^\circ$), the maximum deviation between the ellipse and the approximated circle is below $1\,\mathrm{mm}$ (Fig.~2(b)).
Compared with ~\cite{kim2018quaternion}, our design reduces the ratio $h_c/w_c$ from $80.33/30 \approx 2.68$ to $47/35 \approx 1.34$, corresponding to a reduction of about $50\%$, thereby decreasing the occupied area and inertia.

As shown in Fig.~2(b), two anti-parallelogram structures are arranged orthogonally to form a 2-DOF wrist, thereby enabling omnidirectional rotation. The joint is actuated by four tendons, arranged in opposing pairs. Each pair is driven by one motor, and the tendons are routed through guides to a spool, thereby shifting actuation away from the wrist and reducing distal inertia. In this way, the posteriorly placed motors can effectively drive the parallel anti-parallelogram tendon-driven wrist at the robot’s distal end. To enable accurate sensing, two decoupled joint-angle encoders are mounted on the orthogonal subchains, providing independent roll–pitch feedback without cross-coupling. This compact integration preserves the wrist’s low inertia while supporting high-dynamic control.

\subsection{Kinematics Modeling}

To reflect the axial spinning nature of the handkerchief task, the wrist motion is parameterized by two variables: the rotation-plane angle $\alpha$ and the deflection angle $\beta$. The wrist is approximated as a sphere rolling-contact joint.

\subsubsection{Joint space to Cartesian space}

The forward kinematics from joint space to Cartesian space is derived by sequentially applying the following steps: (i) rotate by $\alpha$ to select the bending plane; (ii) rotate by $\beta$ within the bending plane, established according to the coordinate system (Fig.~2(c)). The homogeneous transformation can thus be expressed as:
\begin{equation}
\begin{aligned}
^oH_e &= R_z(\alpha)R_y\!\left(\tfrac{\beta}{2}\right)T_z(D)R_y\!\left(\tfrac{\beta}{2}\right)R_z(-\alpha) \\
&= \scalebox{0.65}{$
\begin{bmatrix}
 1-2C^2(\alpha)S^2(\tfrac{\beta}{2}) & -2S(2\alpha)S^2(\tfrac{\beta}{2}) & C(\alpha)S(\beta) & DC(\alpha)S(\tfrac{\beta}{2}) \\
 -2S(2\alpha)S^2(\tfrac{\beta}{2}) & 1-2S^2(\alpha)S^2(\tfrac{\beta}{2}) & S(\alpha)S(\beta) & DS(\alpha)S(\tfrac{\beta}{2}) \\
 -C(\alpha)S(\beta) & -S(\alpha)S(\beta) & C(\beta) & D C(\beta) \\
 0 & 0 & 0 & 1
\end{bmatrix}
$}
\end{aligned}
\end{equation}
Here $R_z(\cdot)$ and $R_y(\cdot)$ denote the homogeneous rotation matrices about the $z$ and $y$ axes; $T_z(\cdot)$ denotes the homogeneous translation matrix along the $z$-axis; $C(\cdot)$ and $S(\cdot)$ are abbreviations for the cosine and sine functions; and $D$ represents the diameter of the sphere rolling-contact joint.

\subsubsection{Actuation space to joint space}

Since the wrist fixes the driving tendons at symmetric points on the plane passing through the sphere's center, tendon elongation and contraction remain equal. This symmetry allows pairs of tendons to be driven precisely by a single motor. Meanwhile, with two orthogonal tendon–motor pairs, the joint achieves independent roll and pitch motions in a fully decoupled manner.

Using geometric analysis, the tendon length variations under torsion $\alpha$ and joint rotation $\beta$ can be expressed as
\begin{equation}
\begin{aligned}
l_r &= w \sin(\theta - \alpha)\,\sin\beta, \\
l_p &= w \cos(\theta - \alpha)\,\sin\beta ,
\end{aligned}
\end{equation}
where $\theta$ denotes the angle between the $x$-axis and the tendon tunnel on the joint base, set to $45^\circ$ in the present design.

In summary, this kinematic formulation reveals a clear mapping from tendon displacements to joint rotations through the torsion–bending parameters $(\alpha,\beta)$, and subsequently to Cartesian motions. For the handkerchief-spinning task, such modeling is advantageous because the deflection angle $\beta$ directly corresponds to the spin amplitude, while the rotation-plane angle $\alpha$ corresponds to the spinning axis, enabling intuitive strategy design for periodic control.

\subsection{High–low Level Hierarchical Control for Handkerchief Spinning}

\begin{figure}[!t]
\centering
\includegraphics[width=0.48\textwidth]{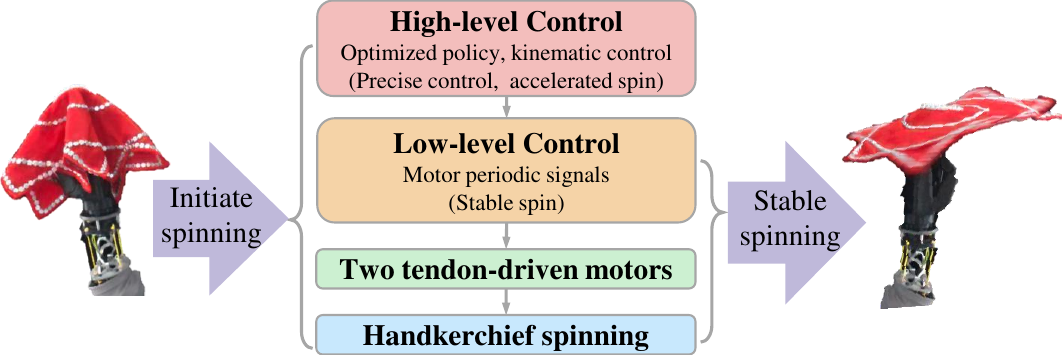}
\caption{High–low level hierarchical control for the handkerchief spin task. The high-level control provides optimized policy and kinematic commands to initiate and regulate spin. The low-level control converts these commands into periodic motor signals that coordinate the two tendon-driven motors, sustaining stable handkerchief spinning.}
\label{3}
\end{figure}

As shown in Fig.~\ref{3}, we establish a hierarchical control framework tailored to the handkerchief-spinning task.  
At the high-level control, the controller plans the initiation and regulates spin using optimized policies and kinematic-based feedback, enabling precise acceleration into periodic motion.  
At the low-level control, the controller transforms these commands into motor-level periodic signals that directly drive the two tendon-actuated motors.  

During the transient phase, the high-level control ensures reliable initiation and acceleration. Once steady spinning is established, the low-level control maintains the periodic rotation with simple signals, improving stability and efficiency.  
This hierarchy unifies precise high-level planning with intuitive low-level stabilization, enabling robust rest-to–steady-state transitions and sustained handkerchief spinning.

\section{Control-Oriented Modeling}
To study how to initiate handkerchief spinning, we develop a particle–spring model that captures the essential dynamics under boundary constraints and frictional contact. This model provides a simulation environment for a serial robotic arm to perform the task and enables systematic evaluation and optimization of initiation strategies.

\subsection{Physical Engine Design}
The simulation system consists of three fundamental physical entities: \textbf{\textit{particles, anisotropic spring constraints, and collision surfaces}}.

\textbf{\textit{The particles}} possess mass $m_i$, position vector $\mathbf{r}_i\in \mathbb{R}^3$, and velocity vector $\mathbf{v}_i\in \mathbb{R}^3$. Some particles exhibit driving attributes (their position and velocity are controlled by external functions), while the remainder are driven particles whose physical states are determined by the system's force transmission and constraint conditions.

\textbf{\textit{The anisotropic spring chain}} serves as a constraint element connecting particles $i$ and $j$, adhering to modified Hooke's law. It exhibits distinct stiffness coefficients for tension $k_s$ and compression $k_c$, relative to its natural length $L_{max,i,j}$, simulating fabric's tendency to fold easily while resisting elongation. Additionally, to ensure simulation convergence, spring damping coefficient $c$ is introduced in the spring chain system to model energy loss during elastic deformation of flexible fabric. Velocity damping $\xi$ is incorporated in the simulation world system to simulate real-world phenomena like air friction.

\textbf{\textit{The collision surface}} consists of a deformable rigid triangular plane formed by three particles, used for collision detection between particles and the fabric surface represented by the triangular plane. The overall collision dynamics adhere to momentum conservation. Due to the properties of flexible materials, collisions in the simulation involve momentum and energy loss. Therefore, an elasticity coefficient $\varepsilon$ is introduced to control the rebound strength of collisions, and a momentum loss coefficient $\eta$ is introduced to control the proportion of total momentum loss.
These components collectively approximate the elastic and dissipative behavior of flexible fabrics under high dynamic energy input. 

\begin{figure}[!t]
\centering
\includegraphics[width=0.48\textwidth]{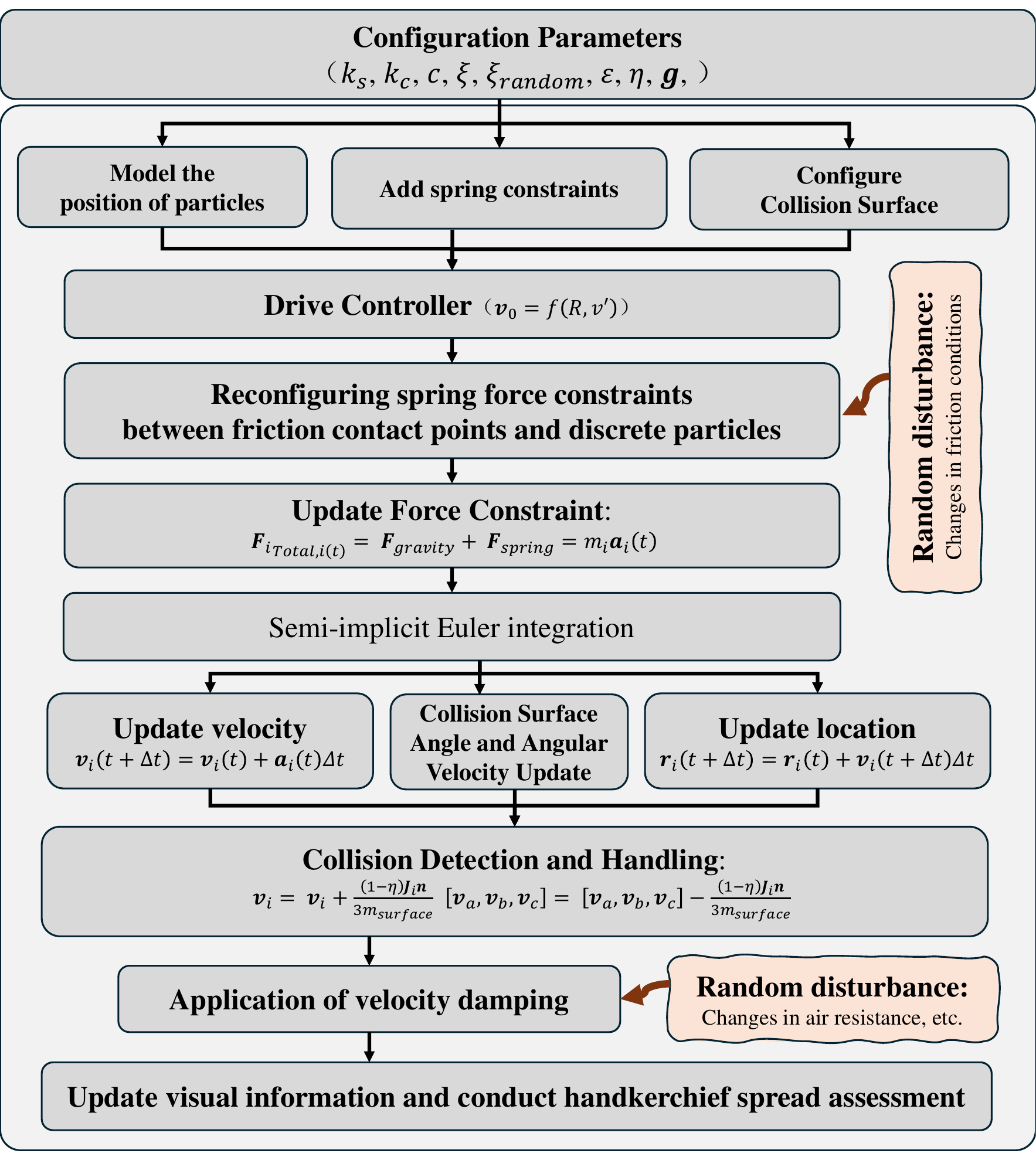}
\caption{Overview of the physical simulation framework for the handkerchief spreading task.}
\label{fig_control_flowchart}
\end{figure}

\subsection{Forces and Dynamics}
During simulation, the total external forces ${{\mathbf{F}}_i}_{Total,\ \ i\left(t\right)}$ acting on particle $i$ include gravity, spring forces, and damping forces. Among these, spring forces simulate the tension transmitted between fabric fibers. Such forces exist not only between dispersed particles but also between the friction contact points of the handkerchief, namely the driving points defined in the simulation, and each individual particle.

Among these, ${\mathbf{g}} = -9.81$ acts on all particles except the driving point, whose motion state is specified by the driving equation. Together, these forces yield a high-dimensional nonlinear system where particle trajectories emerge from coupled elastic, damping, and collisional interactions.

\begin{equation}
k\left(\left\|{\mathbf{r}}_j-{\mathbf{r}}_{i}\right\|-L_{\max , i, j}\right) \frac{{\mathbf{r}}_{i, j}}{\left\|{\mathbf{r}}_{j}-{\mathbf{r}}_{i}\right\|}+c {\mathbf{v}}_{i, j} \frac{{\mathbf{r}}_{i, j} {\mathbf{r}}_{i, j}}{\left\|{\mathbf{r}}_{j}-{\mathbf{r}}_{i}\right\|^{2}}
\end{equation}
where:

\begin{equation}
\boldsymbol{k}=\left\{\begin{array}{l}
k_{s}, if \ L_{c u r, i, j}>L_{\max , i, j}  \\
k_{c}, if \ L_{c u r, i, j}\leqslant L_{\max , i, j}
\end{array}\right.\end{equation}

Here, $L_{\mathrm{cur},i,j}=\left\| \mathbf{r}_j - \mathbf{r}_i \right\|$ is the current spring length, $\mathbf{r}_{i,j}=\mathbf{r}_j-\mathbf{r}_i$, and $\mathbf{v}_{i,j}=\mathbf{v}_j-\mathbf{v}_i$.

Additionally, the impulse method is employed to handle collisions between the particles and triangular surfaces.

\subsection{Driving Point and Boundary Constraints}

In the simulation, in addition to the internal length constraints of the spring chain, a dynamic chain-length constraint is enforced between the driving point and the surrounding particles. The driving point represents the contact location between the finger and the fabric.

Rather than explicitly modeling friction forces, which are difficult to constrain numerically, we directly approximate their effect by assigning a prescribed velocity to the material particle at the contact point along a specified direction. This formulation avoids the instability of frictional modeling while preserving its role in fabric motion initiation.

To capture uncertainties arising from contact posture variations, slippage, and snagging, random perturbations are applied to the contact point position. The geometric relation between the contact point and the particle distribution is defined with respect to the fully unfolded configuration. The contact motion is parameterized as a circular trajectory of radius $R$ centered at the fabric center with angular speed $v'$.

\subsection{Numerical Integration and Damping}
During simulation under gravitational acceleration $\mathbf{g}$, the time step is denoted as $\Delta t$, and each frame is divided into $N$ sub-steps. Particle dynamics are updated using a semi-implicit Euler scheme, which ensures stable trajectory convergence while maintaining sufficient fidelity to capture fabric dynamics.

At each sub-step, the driving point velocity and position are updated according to the driving equation $\mathbf{v}_0 = f(R(t), \mathbf{v}'(t))$, with additional random perturbations. The relative distances between particles and friction-driving points are then computed, and the spring-chain constraints between the driving point and distributed particles are enforced. Particle velocities are updated within the time step under gravity $\mathbf{g}$. 

After force updates, collision handling between particles and fabric triangular faces is performed to obtain corrected velocities. A damping operation is subsequently applied to improve numerical stability and convergence. Velocity damping follows an exponential decay model $\mathbf{v}_i(t+\Delta t) = \xi \cdot \xi_{\text{rand}} \mathbf{v}_i$, where $\xi_{\text{rand}}$ is a stochastic perturbation factor that approximates unpredictable variations such as air resistance during fabric motion. The damped velocities are finally integrated to update particle positions for visualization.

\section{Control Strategies for Initiation}
The handkerchief-spinning task can be formulated as the construction of a periodic steady-state motion. In the model introduced in Sec.~2, this objective is simplified by parameterizing the driving function through time-modulated rotational speed $\alpha(t)$ and rotational radius $\beta(t)$. The resulting driving signal imposes velocity inputs on designated particles corresponding to frictional contact points in the handkerchief model.
This section presents a quantitative metric to evaluate the unfolding progress and analyzes several intuitive initialization strategies for achieving stable periodic motion.
\subsection{Handkerchief Model and Evaluation Function}

\begin{figure}[!t]
\centering
\includegraphics[width=0.5\textwidth]{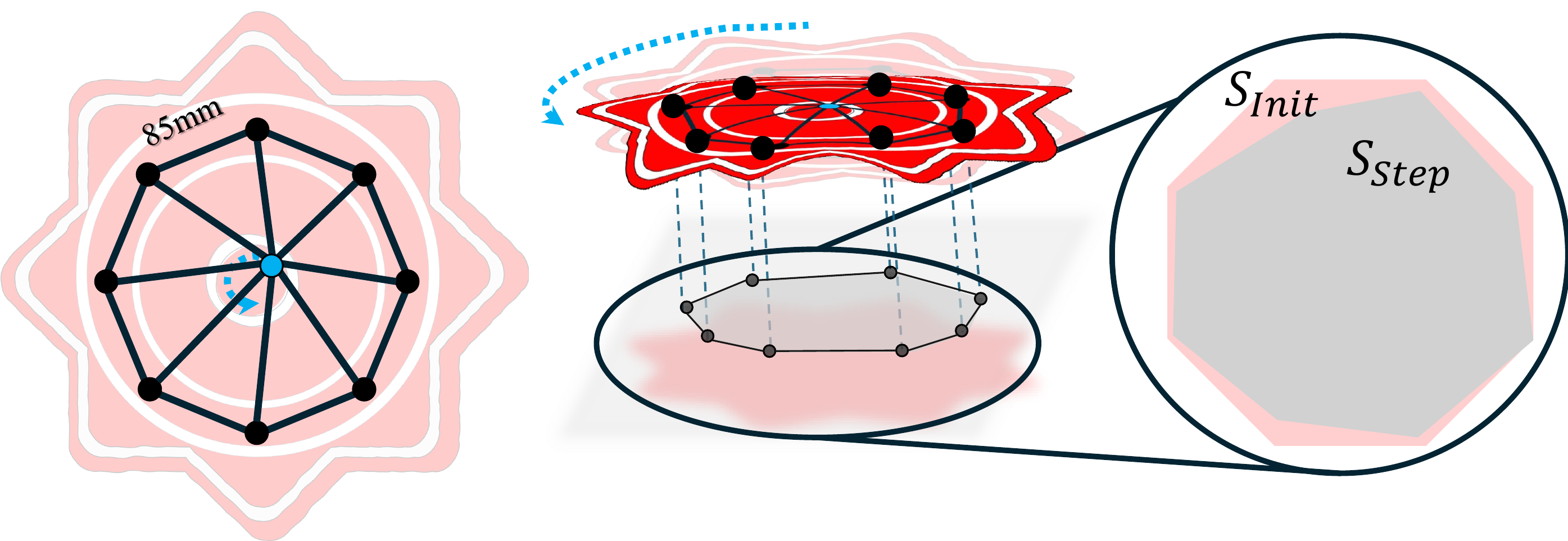}
\caption{Modeling the handkerchief and constructing the degree of handkerchief unfolding: evaluating the unfolding degree of a handkerchief by comparing the projected area on a rotational plane to its fully unfolded area.}
\label{modelling and evaluation}
\end{figure}

Among commercially available performance handkerchiefs, the most common configuration is an octagonal star pattern formed by two squares overlapped at a $45^\circ$ angle. Accordingly, to model high-dynamic tossing behavior, the handkerchief is represented as an octagonal mesh. The eight particles correspond to the centers of mass of eight identical fabric units, assuming uniform mass distribution. The edge-based spring chains, together with the springs connecting the particles to the friction contact points, transmit both dynamic forces and geometric constraints throughout the network.

During high-speed rotational motion, centrifugal effects drive the handkerchief toward planar expansion. The particle positions are projected onto the rotational plane, and the unfolding degree is quantified as the ratio between the area enclosed by these projected particles and the reference area corresponding to the fully unfolded configuration. The projected polygon area is computed using the shoelace formula applied to the planar particle coordinates.

\subsection{Nonlinear Analysis}

Analysis reveals that external conditions and internal constraints are relatively stable. In the dissipative system of handkerchief spinning (involving damping and inelastic collisions), we observe that the relative steady state after the handkerchief begins to spin exhibits periodicity, stability to perturbations, and attractiveness. The Poincaré map provides an effective approach for analyzing periodic steady states and chaotic behavior in flexible systems\cite{ban2022dynamic}. To construct this periodic steady state of handkerchief motion, we employ Poincaré mapping. In the dynamical model of the discrete particle–spring system, we consider the driving equation in the following form:

In the dynamical model of a discrete particle–spring system, we consider the driving equation in the following form:
\begin{equation}
M \ddot{X}(t) + D \dot{X}(t) + K(X(t)) = f(R(t), V(t))
\end{equation}
Here, $M$ denotes the mass matrix, $D$ represents the damping matrix, $K(\cdot)$ signifies the nonlinear spring force term, and $f(R(t), V(t))$ denotes the external input function. Its temporal characteristics directly determine whether the system can achieve and maintain periodic steady-state motion.
Let $T$ denote the fundamental period of the input function. The Poincaré map is defined as $P: \; X(t_0) \mapsto X(t_0+T)$.
Here, $X(t)$ denotes the state of the system in phase space. If there exists a fixed point $X^\ast$ such that $P(X^\ast) = X^\ast$,
then the system possesses a periodic solution with a period coinciding with the input period $T$. Furthermore, if all eigenvalues of the Jacobian matrix of $P$ near $X^\ast$ satisfy $|\lambda_i| < 1$,
the cycle solution is asymptotically stable under the Poincaré map. To ensure discrete particles maintain stable periodic motion, the input function $f(R(t), V(t))$ should satisfy the following temporal characteristics:

\begin{enumerate}
  \item \textbf{Cycle consistency:}  
  The input function must have a well-defined fundamental period $T$, that is,
  \begin{equation}
  f(R(t+T), v(t+T)) = f(R(t), V(t)) \quad \forall t
  \end{equation}
  This ensures the formation of closed orbits in the Poincaré section and guarantees fixed points in the state space.
  \item \textbf{Energy balance:}  
  Within an input cycle of period $T$, the work done by external forces should compensate for energy dissipation, i.e.,
  \begin{equation}
  \int_0^T f(R(t), V(t)) \cdot \dot{X}(t) \, dt 
  \approx \int_0^T D \dot{X}(t) \cdot \dot{X}(t) \, dt.
  \end{equation}
   Otherwise, motion decays due to insufficient energy or diverges due to excess energy.

  \item \textbf{Phase locking:}  
  The input signal should remain phase-locked with the dominant mode of the response to prevent detuning. That is, the input phase $\phi_f(t)$ and the response phase $\phi_x(t)$ satisfy $\phi_f(t) - \phi_x(t) \approx \text{constant}$.

  \item \textbf{Harmonic control:}  
  If the input contains higher-order harmonics, their frequencies should be integer multiples of the fundamental:
  \begin{equation}
  f(R(t), V(t)) = \sum_{n=1}^{\infty} F_n \cos(n \Omega t + \varphi_n), 
  \quad \Omega = \frac{2\pi}{T}.
  \end{equation}
  Otherwise, quasi-periodic or chaotic responses may appear and hinder steady-state maintenance.
\end{enumerate}

To drive the discrete particle system toward a stable periodic motion aligned with the driving cycle, we implement a driving input in which both amplitude and velocity increase linearly over time:
\begin{equation}\left\{\begin{array}{l}
R(t)=R_0 + (R_T - R_0)\frac{\min(t,\ T_k)} {T_k} \\
V(t)=V_0 + (V_T - V_0)\frac{\min(t,\ T_k)} {T_k} \\
\end{array}\right.\end{equation}
Here, $T_k$ represents the rise time scale for rotation radius and speed. It can be defined by the number of revolutions required to reach the target driving state.

\begin{figure}[!t]
\centering
\includegraphics[width=0.5\textwidth]{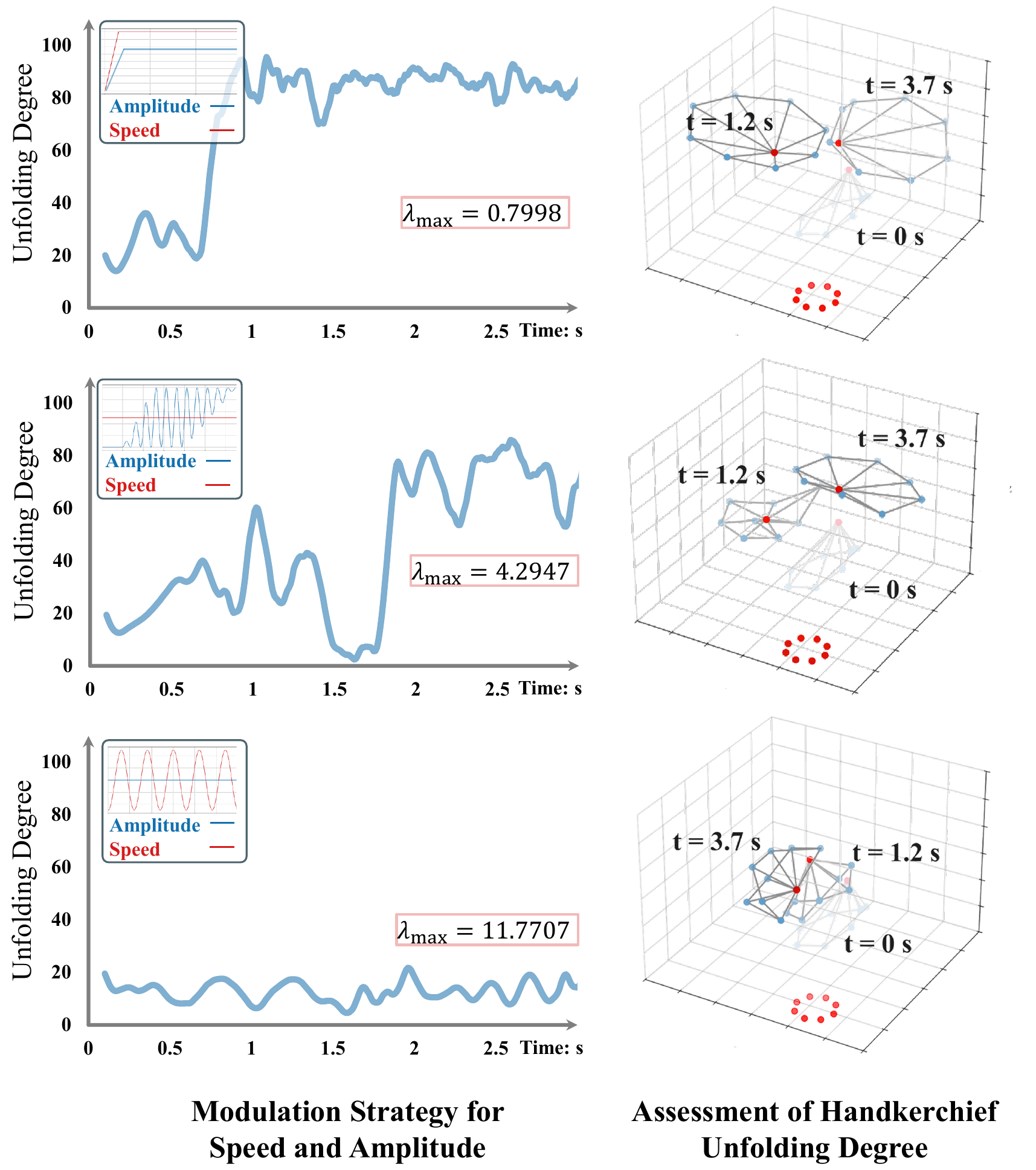}
\caption{Evaluate the unfolding extent of the handkerchief over time under different strategies within the model, with partial time interfaces visualized.}
\label{Comparison of Robustness and Unfolding Degree Across Different Strategies}
\end{figure}

\begin{figure*}[!h]
\centering
\includegraphics[width=1.0\textwidth]{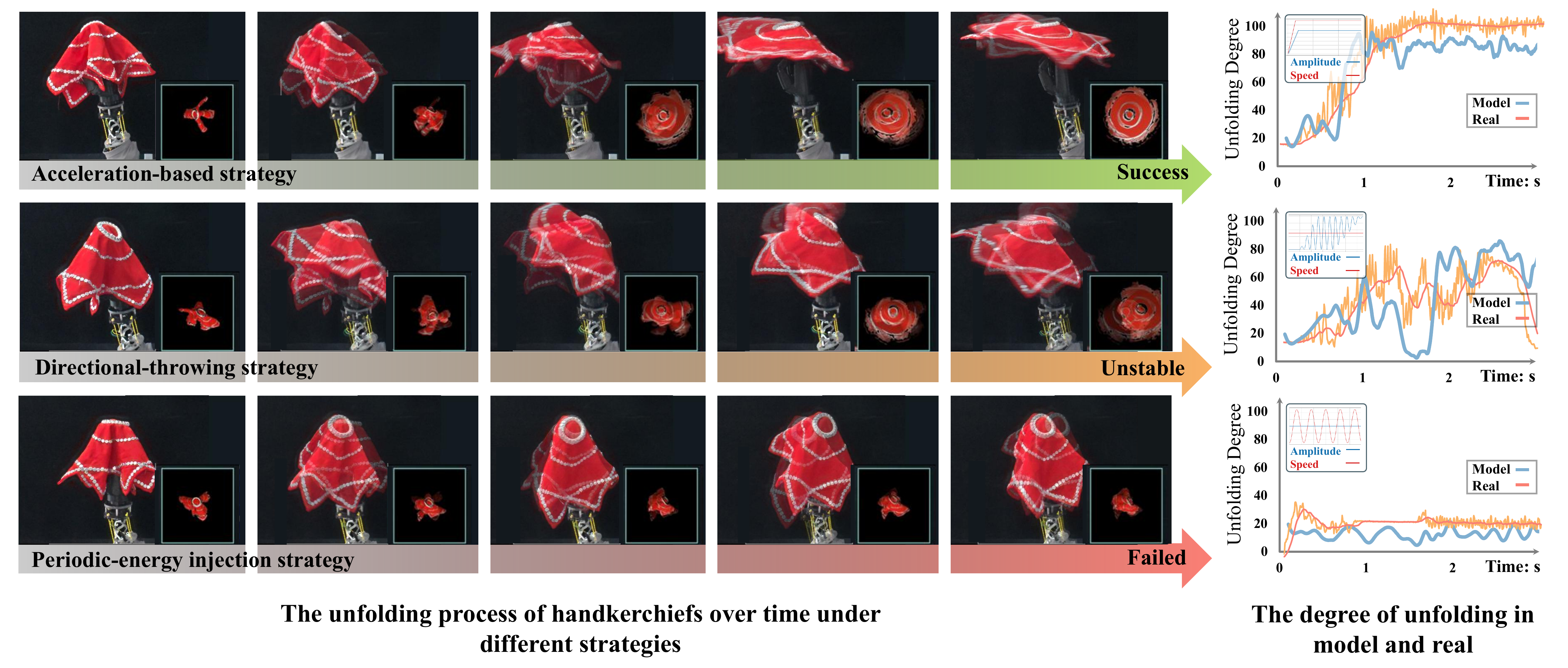}
\caption{Evaluate the unfolding extent of the handkerchief over time under different strategies within the model, with partial time interfaces visualized.}
\label{Experiments and Model Comparisons on Handkerchief unfolding Strategy}
\end{figure*}

The rationale of this strategy lies in three principal considerations. First, a smooth ramp provides quasi-adiabatic traction that limits excitation of higher modes and guides the state toward the desired attractor. Second, the final single dominant frequency couples strongly to the main mode and suppresses unnecessary harmonics. Third, by shaping $R(t)$, the net energy injection per cycle $W_{\mathrm{in}}=\int_{t}^{t+T} f(s)\cdot\dot X(s)\,ds$ can be tuned to balance dissipation and maintain steady amplitude.

Simulation results show that the Poincaré sequence $\{X_n\}$ sampled at $t=nT$ exhibits convergence. By numerically integrating the differential equation along the last period, the single-period fundamental matrix $\Phi(T)$ is obtained, whose eigenvalues (Floquet multipliers) are all less than 1 (maximum magnitude approximately $0.799817$). Meanwhile, the area-based unfolding metric exceeds $90\%$.

Therefore, the periodic orbits are linearly asymptotically stable in the Poincaré sense. Small perturbations diminish with each cycle. Based on the linearization criterion for the Poincaré map, the periodic orbit remains locally asymptotically stable under the given driving.

\subsection{Strategy Analysis and Comparison}
In the previous section, dynamic analysis demonstrated that the acceleration-based strategy for amplitude and speed enables the handkerchief model to achieve periodic steady-state behavior. In dynamic tasks targeting handkerchief spinning, forced motion ultimately leads to a steady-state oscillation synchronized with the driving cycle. However, the process of establishing this steady state depends strongly on the initial condition. The speed and stability of the initial spin reflect the robustness of the transition from rest to a stable periodic state. To identify robust initiation strategies, we compare several alternatives that satisfy the conditions outlined above.

The comparison shows that human-inspired modulation strategies behave differently in the model. Under identical parameters, the acceleration-based strategy increases the radius linearly before holding it constant. The speed also rises linearly before stabilizing. This strategy reaches a periodic steady state within three seconds and fully unfolds the handkerchief. In contrast, the directional-throw strategy uses an elliptic radius transition from a small to a large circle, followed by constant speed. The periodic-energy injection strategy maintains a fixed radius while modulating speed sinusoidally. Neither of these two strategies achieves complete unfolding within the same timeframe. 

Furthermore, Poincaré analysis shows maximum Floquet multipliers of 4.2947 and 11.7707 for the directional-throw and periodic-energy injection strategies. Both values are much greater than one and higher than the 0.7998 observed for the acceleration-based strategy. These results demonstrate the robustness advantage of the acceleration-based modulation. This finding will be further validated in subsequent simulations and real-world experiments.

\section{Experimental verification}
This section presents hardware experiments that validate both the proposed initiation strategies and the hierarchical control framework. We first evaluate different spinning strategies on the developed wrist prototype, and then demonstrate the feasibility of the high–low level control scheme in real-world handkerchief spinning.

\subsection{Performance of Different Strategies on Dexterous Wrist}

To quantify unfolded area of the handkerchief, we employ SAM-V2 to segment the handkerchief mask from images. The median depth $Z$ within the segmented region is extracted and combined with the camera intrinsic parameters to convert the pixel count into an estimated physical area, computed as $A \approx N_{\mathrm{px}} \cdot (Z/f_x) \cdot (Z/f_y)$~(cm$^2$). This formulation provides a real-time estimate of the unfolded area.

As shown in Fig.~7, the experimental results are consistent with the simulation analysis. The acceleration-based strategy, in which both angular speed and rotation radius are gradually ramped to their target values, reliably drives the handkerchief from rest to a fully unfolded steady-state spin within a few seconds. In contrast, oscillatory modulation or large-amplitude directional swings frequently result in incomplete unfolding or unstable transients, indicating reduced robustness under real-world conditions.

Quantitatively, the acceleration-based strategy achieves an unfolding degree of approximately $99\%$ with minimal fluctuation ($CV \approx 0.7\%$). The directional-throw strategy reaches a maximum unfolding of $77.5\%$, but the degree later decreases as trailing fabric catches up with the leading portion and forms folds. The periodic-energy injection strategy remains ineffective, with an average unfolding of only $22.91\%$ throughout the process.  This shows that under the current parameters, adding high-frequency components outside the fundamental period does not improve stability.

Comparison between model predictions and experimental measurements shows close agreement. The model effectively captures the kinematic transition from rest to steady-state spinning under boundary-constrained frictional driving. These results confirm that the control-oriented abstraction, though simplified, is sufficient to guide robust strategy design for high-dynamic fabric manipulation.

\subsection{High–low Level Control of the Dexterous Wrist}

To validate the proposed hierarchical control framework (Fig.~3), we conducted experiments on the tendon-driven wrist under variable-axis spinning tasks. 
In these tests (Fig.~8), the high-level controller specified the spinning axis and initiation strategy, while the low-level tendon actuation executed the commanded trajectories and stabilized transient dynamics. 
The high-level command followed a rotational profile with an angular velocity of $\dot{\alpha}=0.1\,\text{rad/s}$ and a total rotation angle of $\beta=12^\circ$. 
The commanded trajectory in joint space and the corresponding $x$- and $y$-axis profiles demonstrate that the hierarchical control achieves accurate trajectory tracking and coordinated spinning. 

The resulting tracking error of fingertip was relatively small, with $\mathrm{RMSE}=2.88 \text{mm}$ and $R^2=0.9898$, confirming that the system can realize precise, high-dynamic rotational motion around arbitrary axes. 
The residual error is mainly attributed to the inertia of the hand structure during rapid spinning. 
Overall, these results show that the proposed decoupled high-low level hierarchical controller enables coupled control of axis reorientation and handkerchief spinning.

\begin{figure}[!t]
\centering
\includegraphics[width=0.48\textwidth]{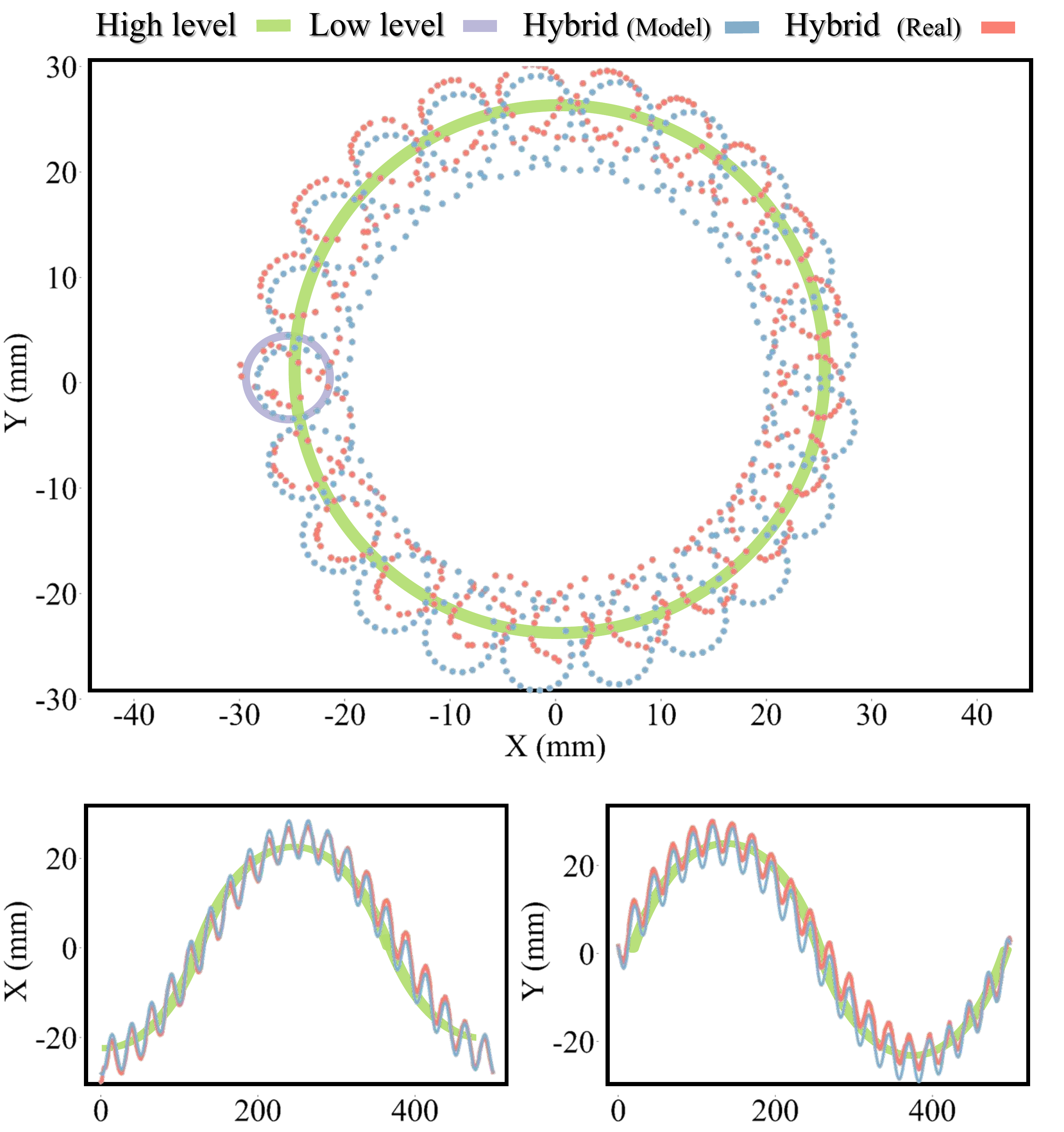}
\caption{Comparison of modeled and experimental trajectories with tracking errors under variable-axis spinning. 
The Cartesian trajectory and $x$/$y$ profiles demonstrate effective hierarchical control, enabling precise axis reorientation and stable handkerchief spinning.}
\label{8}
\end{figure}

\section{Conclusion} 

We developed a parallel anti-parallelogram tendon-driven wrist and validated it through hardware experiments with a high–low level control scheme. The wrist achieves low effective inertia and $90^\circ$ omnidirectional rotation with decoupled roll–pitch sensing, enabling responsive and precise dynamic motion. Experiments demonstrate robust handkerchief spinning with an unfolding ratio of $\approx99\%$ and fingertip tracking error of $\mathrm{RMSE}=2.88\,\text{mm}$, confirming the effectiveness of the proposed architecture. 

Complementing the hardware, we introduced a control-oriented particle–spring model that abstracts friction into a tractable form for controller design and systematic strategy evaluation. The model confirms that a linear ramp of radius and speed reliably drives the cloth from rest to a periodic steady state, consistent with the hardware observations. These findings show that low-inertia wrist hardware, combined with modeling-guided control strategies, provides a practical recipe for robust rest-to–steady-state transitions and accurate periodic manipulation of highly flexible objects. 

Future work will build on this framework to explore learning-based methods for more complex handkerchief-spinning skills and to integrate the bionic wrist into humanoid dual-arm systems for advanced multi-arm coordination, which may further enhance robotic capabilities in handling flexible objects under dynamic conditions.

\section{Acknowledgments}
This work was done at Tsinghua University and was supported by the Joint Funds of the National Key Research and Development Program of China (No. 2024YFB4711102), the National Natural Science Foundation of China (No. U22A2057), the Nanjing Major Science and Technology Special Project (No. 202405017), the Fundamental Research Funds for the Central Universities (No. KG202514), and the National Natural Science Foundation of China (No. 82090053 and 82090050).



\bibliographystyle{IEEEtran}
\bibliography{references.bib}
\end{document}